\pdfoutput=1

\documentclass[11pt]{article}

\usepackage{ACL2023}

\usepackage{times}
\usepackage{latexsym}

\usepackage[T1]{fontenc}

\usepackage[utf8]{inputenc}

\usepackage{microtype}

\usepackage{inconsolata}

\usepackage{diagbox}
\usepackage{multirow}
\usepackage{graphicx}
\usepackage[fleqn]{amsmath}
\usepackage{color}
\usepackage{textcomp}
\usepackage{stfloats}
\usepackage{subfigure}
\usepackage{algpseudocode}
\usepackage{hyperref}
\usepackage[linesnumbered,ruled]{algorithm2e}

\title{Prompt Tuning Pushes Farther, Contrastive Learning Pulls Closer:\\ A Two-Stage Approach to Mitigate Social Biases}

\author{Yingji Li\textsuperscript{1}, 
Mengnan Du\textsuperscript{2}, Xin Wang\textsuperscript{3}\thanks{~~Corresponding author}~~, 
Ying Wang\textsuperscript{1,4}\footnotemark[1]\\
\textsuperscript{1}College of Computer Science and Technology, Jilin University, Changchun, China \\
\textsuperscript{2}Department of Data Science, New Jersey Institute of Technology, Newark, USA \\
\textsuperscript{3}School of Artificial Intelligence, Jilin University, Changchun, China \\
\textsuperscript{4}Key Laboratory of Symbolic Computation and Knowledge Engineering\\of Ministry of Education, Jilin University, Changchun, China  \\
\small\texttt{yingji21@mails.jlu.edu.cn}, \small\texttt{mengnan.du@njit.edu},
\small\texttt{\{xinwang,wangying2010\}@jlu.edu.cn}
}

\begin{document}
\maketitle
\begin{abstract}
As the representation capability of Pre-trained Language Models (PLMs) improve, there is growing concern that they will inherit social biases from unprocessed corpora. Most previous debiasing techniques used Counterfactual Data Augmentation (CDA) to balance the training corpus. However, CDA slightly modifies the original corpus, limiting the representation distance between different demographic groups to a narrow range. As a result, the debiasing model easily fits the differences between counterfactual pairs, which affects its debiasing performance with limited text resources. In this paper, we propose an adversarial training-inspired two-stage debiasing model using \underline{C}ontrastive learning with \underline{C}ontinuous \underline{P}rompt \underline{A}ugmentation (named CCPA) to mitigate social biases in PLMs' encoding. In the first stage, we propose a data augmentation method based on continuous prompt tuning to push farther the representation distance between sample pairs along different demographic groups. In the second stage, we utilize contrastive learning to pull closer the representation distance between the augmented sample pairs and then fine-tune PLMs' parameters to get debiased encoding. Our approach guides the model to achieve stronger debiasing performance by adding difficulty to the training process. Extensive experiments show that CCPA outperforms baselines in terms of debiasing performance. Meanwhile, experimental results on the GLUE benchmark show that CCPA retains the language modeling capability of PLMs.
\end{abstract}

\section{Introduction}
\begin{figure}[t]
\vspace{10pt}
    \begin{center}
        \makeatletter\def\@captype{figure}\makeatother
        \includegraphics[width=0.37\textwidth]{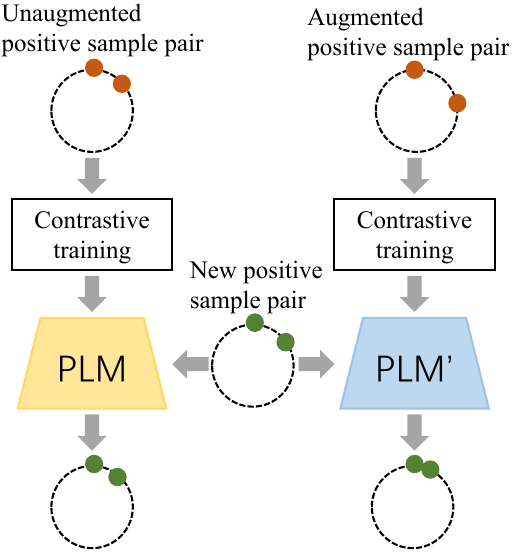}
        \caption{The motivation of CCPA. For the new input samples, the PLM's performance of the augmented sample training is stronger than that of the unaugmented sample training.}
        \vspace{-7pt}
    \label{fig1}
    \end{center}
\end{figure}
Pre-trained Language Models (PLMs) have demonstrated outstanding performance in recent years and have been widely used in natural language understanding tasks \citep{DBLP:conf/naacl/PetersNIGCLZ18,DBLP:conf/naacl/DelobelleTCB22}. However, the powerful language modeling capability enables PLMs to learn good representations from large-scale training corpora while capturing human-like social biases. Recent studies have demonstrated that the representations encoded by PLMs learn social biases specific to demographic groups (e.g., gender, race, religion) and can be amplified to downstream tasks, leading to unfair outcomes and adverse social effects \citep{DBLP:conf/naacl/ZhaoWYCOC19,DBLP:journals/corr/abs-2010-06032}. As a result, mitigating social biases in PLMs' encoding can improve the fairness of NLP systems significantly \citep{DBLP:conf/nips/BolukbasiCZSK16,DBLP:journals/tacl/BenderF18}.

Most existing debiasing techniques first need to construct sample pairs using Counterfactual Data Augmentation (CDA) \citep{DBLP:conf/acl/ZmigrodMWC19,DBLP:conf/acl/0003XSHTGJ22} to balance the training corpora. The general approach of CDA is to replace the original corpus with attribute words (e.g., \textit{he}/\textit{she}, \textit{man}/\textit{woman}) specific to different demographic groups. For example, RCDA \citep{DBLP:conf/emnlp/ChenXY21} uses a generator to generate a large number of antisense sentences and then uses a discriminator to evaluate the quality of the original and antisense samples jointly. FairFil \citep{DBLP:conf/iclr/ChengHYSC21} obtains a pair of positive sample sentences by replacing the attribute words in the training corpora with the antonyms and then uses contrastive learning to train a filter for debiasing. Auto-Debias \citep{DBLP:conf/acl/GuoYA22} uses pairs of attribute words as training corpora, amplifies the bias between sample pairs by searching biased prompt texts in the Wikipedia vocabulary, and then performs semantic alignment using Jensen-Shannon divergence. These methods aim to mitigate social biases between different demographic groups by narrowing the representation distance between sample pairs. 
However, CDA slightly modifies the original corpus, limiting the representation distance between different demographic groups to a narrow range. As a result, the debiasing model is easy to overfit the difference between counterfactual pairs, which affects its learning ability with limited text resources. As shown in Figure~\ref{fig1}, it is difficult for PLMs to achieve the ideal debiasing performance for newly input samples with greater difficulty.

In this work, we propose a two-stage debiasing method using \underline{C}ontrastive learning with \underline{C}ontinuous \underline{P}rompt \underline{A}ugmentation (named CCPA) to mitigate social biases in PLMs' encoding. Inspired by adversarial training, our approach improves the debiasing ability of PLMs by first amplifying and then attenuating the bias between different demographic groups. Specifically, we first use CDA to replace attribute words in the original training corpus to construct counterfactual pairs corresponding to different demographic groups. In the first stage, we augment the positive sample pairs with continuous prompt tuning to increase the distance between them to amplify the biases between different demographic groups. In the second stage, we utilize contrastive learning to pull the distance between the positive sample pairs to attenuate the biases between different demographic groups. CCPA increases the difficulty of model fitting by expanding the representation space between sample pairs. We believe that difficult learning experiences make the model more powerful, thus improving the debiasing ability of PLMs training in corpora with limited resources. Our main contributions are as follows:
\begin{itemize}
    \item 
    We propose the CCPA debiasing framework that combines prompt tuning and contrastive learning to learn a debiased PLM representation. The PLM's parameters are fixed in the first stage, and a generator encoding continuous prompts is trained. In the second stage, the prompts are fixed, and the PLM's parameters are fine-tuned using contrastive learning. 
    \item We propose data augmentation using continuous prompts to achieve excellent debiasing performance using small training data rather than relying on a large external corpus. Given that continuous prompts may cause the representation distance between sample pairs to be too far apart, causing the semantic space to degrade, we propose constraining the prompt tuning using the Mahalanobis Distance to keep the semantic space as stable as possible.
    \item We train CCPA on several real-world corpora and mitigate bias on the most common gender bias. The results on BERT and DistilBERT show that CCPA is superior to state-of-the-art models. In addition, we test the downstream tasks on the GLUE benchmark, and show that CCPA retains the language modeling capability while improving the PLMs' fairness.
\end{itemize}

\begin{figure*}[t]
    \begin{center}
        \makeatletter\def\@captype{figure}\makeatother
        \includegraphics[width=0.8\textwidth]{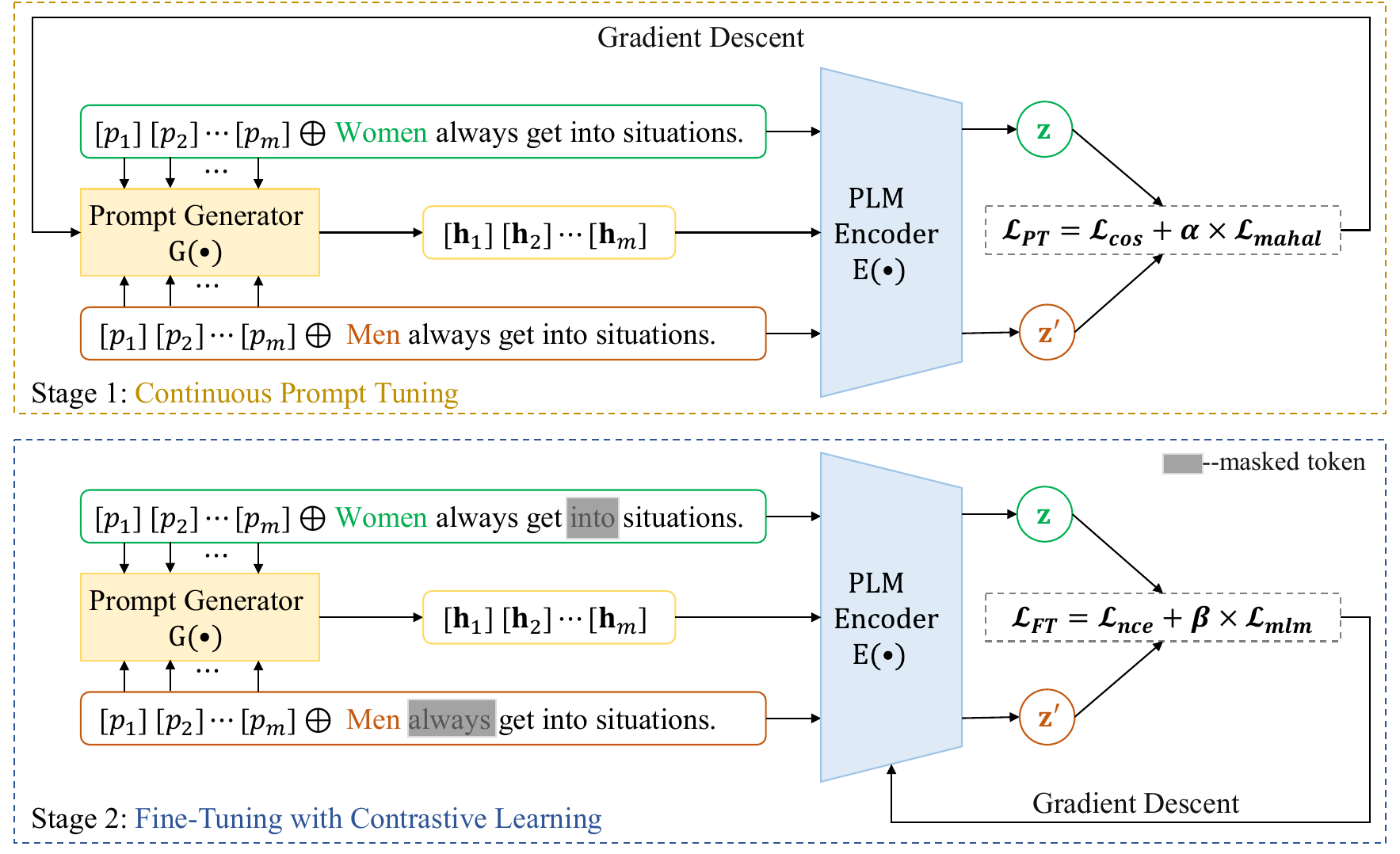}
        \caption{The overall architecture of CCPA. In the first stage, the parameters of PLM encoder $E(\cdot)$ are fixed and a prompt generator $G(\cdot)$ encoding the continuous prompts is trained, where the goal is to enlarge the bias of sentence pairs. In the second stage, the parameters of the prompt generator $G(\cdot)$ are fixed and the parameters of PLM encoder $E(\cdot)$ are fine-tuned using contrastive loss. Ultimately, we can obtain the debiased PLM encoder $E(\cdot)$.}
    \label{fig2}
    \end{center}
\end{figure*}

\section{Methodology}
In this section, we propose the \underline{C}ontrastive learning with \underline{C}ontinuous \underline{P}rompt \underline{A}ugmentation (CCPA) framework to mitigate the social bias in the encoding of PLMs specific to the most common gender bias. Our proposed CCPA consists of two stages: 1) Continuous Prompt Tuning and 2) Fine-Tuning with Contrastive Learning. The framework of CCPA is shown in Figure~\ref{fig2}.

\subsection{Pre-Processing based on CDA}
First, we pre-process the training corpus with imbalanced samples using Counterfactual Data Augmentation (CDA). Given a list of attribute words specific to gender bias,\footnote{We only consider the binary gender direction and use the same list of gender-specific attribute words as  \citep{DBLP:conf/nips/BolukbasiCZSK16,DBLP:conf/acl/LiangLZLSM20,DBLP:conf/iclr/ChengHYSC21}.} for each attribute word (e.g., \textit{male}/\textit{female}), we match sentences containing an attribute word in the training corpus. The attribute word is then replaced with the opposite word in a different gender direction (e.g., \textit{male} is replaced by \textit{female}), leaving the other words unchanged. Then, we get the pre-processed training corpus $\mathcal{S}=\{(s_1,s_1'),(s_2,s_2'),\cdots,(s_N,s_N')\}$ consists of $N$ counterfactual pairs $(s_i,s_i')$ along different gender directions.

\subsection{Continuous Prompt Tuning}
Prompt-based learning is similar to giving instructions to the model task to guide the model learning knowledge more directly \citep{DBLP:conf/emnlp/PetroniRRLBWM19}. A lot of work utilize manually constructed prompts \citep{DBLP:journals/corr/abs-2012-11926,DBLP:conf/eacl/SchickS21} or automatically searched discrete prompts \citep{DBLP:conf/emnlp/ShinRLWS20} to assist language models. However, manually constructed templates are heavily based on the designers' experience and automatically searched prompts are limited by the search space \citep{DBLP:journals/corr/abs-2107-13586}. Instead of limiting the prompts to human interpretable natural language, the continuous prompts \citep{DBLP:conf/acl/LiL20,DBLP:conf/naacl/ZhongFC21} guide directly within the embedding space of the model. Meanwhile, continuous prompts tune their parameters, removing the constraint of templates being parameterized by PLMs' parameters.

Inspired by adversarial training, we believe that increasing the difficulty of the training process can guide the model in acquiring a stronger learning ability. To achieve this goal, we propose a data augmentation method based on continuous prompt tuning to further push the differences between counterfactual pairs. Data augmentation method based on continuous prompt tuning adds difficult information to the model by concatenating embeddings that amplify bias across different demographic groups over counterfactual pairs.

Given a template $T=\{[p_1],[p_2],\cdots,[p_{m}],\\s\}$, where $s$ denotes a sentence, $[p_j]$ is a virtual token represented as $[PROMPT]$ and $m$ virtual tokens form a prompt sequence $\mathcal{P}$. For each counterfactual pair $(s_i,s_i')\in\mathcal{S}$ obtained by data pre-processing, we concatenate the same prompt sequence $\mathcal{P}$ at the head of each sentence (see Figure~\ref{fig2}). The augmented sample pair is denoted by $(\hat{s_i},\hat{s_i}')$ and is fed into a PLM to obtain the sentence representation. Formally, let $M$ denote a PLM whose encoder $E(\cdot)$ encodes an input sentence $\hat{s_i}$ and outputs a sentence embedding $\mathbf{z}_i=E(\hat{s_i})$. Similarly, $\mathbf{z}'_i=E(\hat{s_i}')$. In order to obtain continuous prompt embeddings, we train a generator $G(\cdot)$ to encode the prompt sequence $\mathcal{P}$. Following P-Tuning \citep{DBLP:journals/corr/abs-2103-10385}, we choose a bidirectional long-short-term memory network (LSTM), which consists of a two-layer multilayer perceptron (MLP) and a ReLU activation layer. The embedding $\mathbf{h}_j$ of each virtual token $[p_j]$ in the prompts sequence is encoded by $G(\cdot)$ as  follows:
\begin{equation}
    \begin{split}
    \mathbf{h}_j&=G([\overrightarrow{\mathbf{h}_j}:\overleftarrow{\mathbf{h}_j}]) \\
    &=G([LSTM(\mathbf{h}_{1:j}):LSTM(\mathbf{h}_{j:m+1})]).
    \end{split}
\end{equation}
Afterwards, we replace the continuous prompt embeddings $\{\mathbf{h}_1,\mathbf{h}_2,\cdots,\mathbf{h}_{m}\}$ to the corresponding positions of the sentence embeddings $\mathbf{z}_i$ to obtain the sentence representations pairs $(\mathbf{z}_i,\mathbf{z}_i')$.

In this stage, our training objective is to push away the distance of representation $(\mathbf{z}_i,\mathbf{z}_i')$ between sample pairs $(\hat{s_i},\hat{s_i}')$. Briefly, we take the Cosine Similarity between sentence representations as the loss function, defined as follows:
\begin{equation}
    \begin{split}
    \mathcal{L}_{cos}=\frac{\mathbf{z}\cdot \mathbf{z}'}{\|\mathbf{z}\| \|\mathbf{z}'\|} 
    =\frac{\sum_{i=1}^{n}\mathbf{z}_i\cdot \mathbf{z}_i'}{\sqrt{\sum_{i=1}^{n}\mathbf{z}_i^2}\sqrt{\sum_{i=1}^{n}\mathbf{z}_i'^2}},
    \end{split}
    \label{eq2}
\end{equation}
where $\mathbf{z}$ and $\mathbf{z}'$ denote sentence representations with different sensitive attributes within a batch of size $n$, respectively. The representation distance between the sample pairs is enlarged with the gradient of similarity decreasing, thus amplifying the bias information between different genders. 

Considering that the sentence representation with high-dimensional linear distribution is not independently and equally distributed among the dimensions, only relying on Euclidean distance training may cause the sentence representation to deviate from the original distribution and thus destroy the semantic information. To constrain the change of sentence representation within the original distribution, Mahalanobis distance is taken as the regularization term of the loss function:
\begin{equation}
    \mathcal{L}_{mahal}=\sqrt{(\mathbf{z}-\mathbf{S})^{\top}\Sigma^{-1}(\mathbf{z}-\mathbf{S})},
\end{equation}
where $\mathbf{z}$ is the representation of a batch size of samples with concatenated prompt embeddings, $\mathbf{S}$ is the representation of the entire pre-processed training samples without concatenated prompt embeddings, and $\Sigma$ is the covariance matrix of $\mathbf{S}$. Mahalanobis distance is a correction of the Euclidean distance, which corrects the assumption that the Euclidean distance is independent and equally distributed among all dimensions. With the constraint of Mahalanobis distance, the augmented samples of each batch can vary within the distribution range of the original training data to maintain the semantics.

The overall loss function of the continuous prompt tuning stage is defined as:
\begin{equation}
    \mathcal{L}_{PT}=\mathcal{L}_{cos}+\alpha\times\mathcal{L}_{mahal},
    \label{eq_PT}
\end{equation}
where $\alpha$ is a hyperparameter that adjusts the weight of $\mathcal{L}_{mahal}$. In the gradient descent process of $\mathcal{L}_{PT}$, we only adjust the parameters of the generator $G(\cdot)$ and fix the PLMs' parameters to obtain the continuous prompt embeddings that further amplifies the bias between different sensitive attributes.

\subsection{Fine-Tuning with Contrastive Learning}
We then use contrastive learning to mitigate the social bias in PLMs' encoding for different demographic groups. Contrastive learning \citep{DBLP:conf/acl/YangCLS19} is a task-agnostic self-supervision method that learns data features by minimizing contrastive loss to maximize the similarity of the representation vectors of positive sample pairs \citep{DBLP:conf/acl/DasKPZ22}. Specifically, we encourage as much consistency as possible among representations of different sensitive attributes by maximizing the similarity of the augmented counterfactual pairs. Noise Contrast Estimation \citep{DBLP:journals/jmlr/GutmannH10} is usually used as a contrastive loss function, given an augmented sample pair of a batch $\{(\hat{s}_i,\hat{s}_i')\}_{i=1}^n$, which is defined as follows:
\begin{equation}
    \mathcal{L}_{nce}=\frac{1}{n}\sum_{i=1}^n\log\frac{e^{sim(\mathbf{z}_i,\mathbf{z}_i')/ \tau}}{\frac{1}{n}\sum_{j=1}^n e^{sim(\mathbf{z}_i,\mathbf{z}_j)/ \tau}},
    \label{nce}
\end{equation}
where $(\mathbf{z}_i,\mathbf{z}_i')=(E(\hat{s}_i),E(\hat{s}_i'))$, $\tau$ is a temperature hyperparameter and $sim(\cdot,\cdot)$ denotes the similarity function usually using cosine similarity. During training, we only fine-tune the PLMs' parameters and fix the embedding of continuous prompts. By maximizing $\mathcal{L}_{nce}$, differences in the encoding of PLM outputs specific to different demographic groups are eliminated, resulting in representations independent of sensitive attributes.

\begin{algorithm}[tb]
\small
\caption{Proposed CCPA framework.}
\label{alg1}
   \DontPrintSemicolon
   \SetAlgoLined
   \KwIn {Pre-processed training corpus $\mathcal{S}$, PLM encoder $E(\cdot)$, Initial prompt generator $G(\cdot)$, Prompt template $T$, Hyperparameter $\alpha, \beta, \tau$. }
   \While{stage 1}{
   Apply $T$ to $\forall(s_i,s_i')\in\mathcal{S}$ to obtain $(\hat{s}_i,\hat{s}_i')$; \;
   Obtain $(\mathbf{z}_i,\mathbf{z}_i')=(E(\hat{s}_i),E(\hat{s}_i'))$; \;
   Replace $\{\mathbf{h}_1,\mathbf{h}_2,\cdots,\mathbf{h}_{m}\}$ encoded by $G(\cdot)$ in the corresponding position in $(\mathbf{z}_i,\mathbf{z}_i')$; \;
   Calculate $\mathcal{L}_{cos}$ and $\mathcal{L}_{mahal}$ with $\{(\mathbf{z}_i,\mathbf{z}_i')\}_{i=1}^n$; \;
   Update $G(\cdot)$'s parameters following Equation~\ref{eq_PT}; \;
   }
   \While{stage 2}{
   Mask $\forall(s_i,s_i')$ randomly with a 15\% probability;\;
   Obtain $\{(\mathbf{z}_i,\mathbf{z}_i')\}_{i=1}^n$ using $E(\cdot)$ and $G(\cdot)$; \;
   Calculate $\mathcal{L}_{nce}$ and $\mathcal{L}_{mlm}$ and update $E(\cdot)$'s parameters following Equation~\ref{eq_FT}. \;
   }
\end{algorithm}

Considering that the attenuation of biases towards encoding may affect PLMs' language modeling capability, we add a Masking Language Modeling (MLM) loss during the fine-tuning stage to aid PLM training \citep{he2022mabel}. Following previous work \cite{DBLP:conf/naacl/DevlinCLT19}, we randomly mask tokens in training texts with a $15\%$ probability.\footnote{In practice, the chosen masked token has an $80\%$ chance of being masked, a $10\%$ chance of being replaced with another word, and a $10\%$ chance of remaining unchanged.} Our objective is to train the encoder to predict the masked tokens through contextual semantics, thereby preserving the language modeling capability of PLMs.
The overall loss function in the fine-tuning stage is defined as follows:
\begin{equation}
\mathcal{L}_{FT}=\mathcal{L}_{nce}+\beta\times\mathcal{L}_{mlm},
\label{eq_FT}
\end{equation}
where $\beta$ is a hyperparameter that controls the weight of $\mathcal{L}_{mlm}$. 
Our overall algorithm is given in Algorithm~\ref{alg1}.

\section{Experiments}
In this section, we conduct experiments to evaluate the performance of CCPA, in order to answer the following three research questions. 

\textbf{Q1.} How effective is CCPA in mitigating social biases in PLMs' encoding? 

\textbf{Q2.} How does each component affect CCPA?

\textbf{Q3.} Will CCPA preserve the language modeling capability of PLMs?

\subsection{Experimental Setup}
\subsubsection{Attribute Word List \& Datasets}
Following \citep{DBLP:conf/nips/BolukbasiCZSK16,DBLP:conf/acl/LiangLZLSM20,DBLP:conf/iclr/ChengHYSC21,he2022mabel}, our gender attribute word list is set to:

\noindent\{\textit{MALE, FEMALE}\}=\{\textit{(man, woman), (boy, girl), (he, she), (father, mother), (son, daughter), (guy, gal), (male, female), (his, her), (himself, herself), (John, Mary)}\}.

Following \citep{DBLP:conf/acl/LiangLZLSM20,DBLP:conf/iclr/ChengHYSC21}, we select five real-world datasets as the initial training corpus, which are Stanford Sentiment Treebank \citep{DBLP:conf/emnlp/SocherPWCMNP13}, POM \citep{DBLP:conf/icmi/ParkSCSM14}, WikiText-2 \citep{DBLP:conf/iclr/MerityX0S17}, Reddit \citep{DBLP:conf/emnlp/VolskePSS17} and MELD \citep{DBLP:conf/acl/PoriaHMNCM19} respectively. We set the maximum sentence length to 100, and the pre-processed training corpus contained 10,510 sentences.

\subsubsection{Baselines \& Implementation Details}
We select seven recent task-agnostic debiasing models as baselines. \textbf{CDA} \citep{DBLP:conf/acl/ZmigrodMWC19}, \textbf{Dropout} \citep{DBLP:journals/corr/abs-2010-06032}, \textbf{Sent-Debias} \citep{DBLP:conf/acl/LiangLZLSM20}, \textbf{FairFil} \citep{DBLP:conf/iclr/ChengHYSC21}, \textbf{INLP} \citep{DBLP:conf/acl/RavfogelEGTG20} and \textbf{MABEL} \citep{he2022mabel} apply counterfactual data augmentation to sentence-level debiasing, where \textbf{FairFil} and \textbf{MABEL} adopt the contrastive learning framework training model. \textbf{Auto-Debias} \citep{DBLP:conf/acl/GuoYA22} directly uses the attribute word list and the stereotype words list as the training corpus.

We perform the main experiments on BERT \citep{DBLP:conf/naacl/DevlinCLT19} and compare CCPA to all baseline models. We also test debiasing performance on DistilBERT \citep{DBLP:journals/corr/abs-1910-01108} and ELEATRA~\citep{DBLP:conf/iclr/ClarkLLM20}. All checkpoints use \textit{bert-base-uncased}, \textit{distilbert-base-uncased}, and \textit{google/electra-base-generator} implemented by Huggingface Transformers library \cite{DBLP:conf/emnlp/WolfDSCDMCRLFDS20}. In the continuous prompt tuning stage, the learning rate is set to $1e^{-5}$, the batch size is set to 64 and $\alpha=0.005$. Following P-Tuning \citep{DBLP:journals/corr/abs-2103-10385}, the virtual tokens template of continuous prompts is denoted as a triplet with the length of each element selected on $\{1,2,3\}$. In the fine-tuning stage, the learning rate is set to $1e^{-4}$. The batch size is set to 32, $\beta=1$ and $\tau=1$. We report the average of the results of three runs over 20 epochs.

To compare the baseline models more fairly, we apply the same attribute word lists and training datasets to CDA and Dropout as CCPA. The implementation codes for CDA, Dropout, Sent-Debias, and INLP are provided by \citep{DBLP:conf/acl/MeadePR22}, and the implementation codes for FairFil and Auto-Debias are provided by the authors. For MABEL, we report the results from its original paper.

\subsection{Evaluation Metrics}
We measure debiasing performance using the common three internal bias evaluation metrics and two external bias evaluation metrics.

\subsubsection{Internal Bias Evaluation Metrics}
\noindent\textbf{Sentence Encoder Association Test (SEAT)} \citep{DBLP:conf/naacl/MayWBBR19} uses sentence templates to evaluate the association between different sensitive attribute demographic and target concepts. Given the attribute word lists $\mathcal{A}$ and $\mathcal{B}$, the target words lists $\mathcal{X},\mathcal{Y}$. The results are presented by effect size, defined as:
\begin{equation}
\small
    d=\frac{\mu(\{s(x,\mathcal{A},\mathcal{B})\})-\mu(\{s(y,\mathcal{A},\mathcal{B})\})}{\sigma(\{s(t,\mathcal{X},\mathcal{Y})\}_{t\in\mathcal{A}\cup\mathcal{B}})},
    \label{seat}
\end{equation}
where $x\in\mathcal{X}$ and $y\in\mathcal{Y}$, $\mu(\cdot)$ is the mean function and $\sigma(\cdot)$ is the standard deviation. And $s(w,\mathcal{A},\mathcal{B})$ is the bias degree defined as:
$
    s(w,\mathcal{A},\mathcal{B})=\mu(cos(w,a))-\mu(cos(w,b)).
$

The gender-specific subsets of SEAT are 6, 6b, 7, 7b, 8, and 8b. We report the effect size of debiasing models on each subset and the average value of the absolute value of the six subsets, respectively. 

\noindent\textbf{StereoSet} \citep{DBLP:conf/acl/NadeemBR20} uses the fill-in-the-blank template to investigate the stereotype association of PLM. The Language Modeling Score (LM) is the percentage of stereotype or anti-stereotype words selected by the model based on incomplete contextual sentences. The Stereotype Score (SS) is the percentage of models that choose stereotypes over anti-stereotypes. The Idealized Context Association Test (ICAT) is a comprehensive evaluation index of LM and SS.

\noindent\textbf{Crowdsourced Stereotype Pairs (CrowS-Pairs)} \cite{DBLP:conf/emnlp/NangiaVBB20} is a dataset containing pairs of stereotype sentences and anti-stereotype sentences. We report the ratio of mask token probabilities assigned to stereotype sentences rather than anti-stereotype sentences, denoted using CrowS.

\begin{table*}
\centering
\resizebox{\textwidth}{!}{
\begin{tabular}{l|ccccccc|ccc|c}
\hline
\diagbox{Model}{Metric} & SEAT-6 & SEAT-6b & SEAT-7 & SEAT-7b & SEAT-8 & SEAT-8b & Avg. & LM & SS & ICAT & CrowS \\
\hline
\textbf{BERT} & 0.932 & 0.090 & -0.124 & 0.937 & 0.783 & 0.858 & 0.621 & 84.17 & 60.28 & 66.86 & 57.86  \\
+CDA & 0.596 & -0.103 & -0.236 & 0.800 & 0.394 & 0.734 & 0.477 & 85.47 & 58.69 & 70.63 & 55.35 \\
+Dropout & 0.912 & 0.121 & 0.321 & 0.857 & 0.777 & 0.867 & 0.642 & 85.42 & 60.11 & 68.16 & 55.35 \\
+Sent-Debias & 0.336 & -0.314 & -0.624 & \textbf{0.514} & 0.391 & 0.436 & 0.436 & \textbf{85.60} &  59.05 & 70.11 & 42.14 \\
+FairFil & 0.683 & -0.140 & -0.616 & 0.839 & \textbf{0.049} & -0.501 & 0.471 & 48.78 & \textbf{46.44} & 45.31 & 62.89 \\
+Auto-Debias & 0.373 & \textbf{-0.056} & 0.745 & 1.175 & 0.856 & 0.823 & 0.671 & 81.76 & 57.33 & 69.78 & 52.83 \\
+INLP & 0.619 & -0.226 & 0.326 & 0.591 & 0.430 & 0.549 & 0.457 & 82.69 & 58.09 & 69.31 & 50.94 \\
+MABEL & 0.664 & 0.167 & 0.479 & 0.647 & 0.465 & 0.570 & 0.499 & 84.80 & 56.92 & 73.07 & \textbf{50.76} \\
+CCPA (Ours) & \textbf{0.181} &-0.317& \textbf{0.104} & 0.633 &0.142 & \textbf{0.115} &\textbf{0.249} &84.44 &56.61 &\textbf{73.28} & 51.57\\
\hline
\textbf{DistilBERT} & 1.380 & 0.446 & -0.179 & 1.242 & 0.837 & 1.217 & 0.883 & \textbf{84.75} & 60.52 & 66.93 & 59.75  \\
+CCPA (Ours) & \textbf{0.409} & \textbf{-0.024} & \textbf{0.138} & \textbf{-0.029} & \textbf{-0.029} & \textbf{0.283} & \textbf{0.152} & 81.91 & \textbf{56.47} & \textbf{71.30} & \textbf{50.31} \\
\hline
\textbf{ELEATRA} & 0.820 & 0.036 & 1.180 & 1.007 & 0.782 & 0.958 & 0.797 & \textbf{85.12} & 58.15 & 71.24 & 52.83  \\
+CCPA (Ours) & \textbf{0.251} & \textbf{0.012} & \textbf{0.647} & \textbf{0.437} & \textbf{0.720} & \textbf{0.460} & \textbf{0.421} & 84.63 & \textbf{52.97} & \textbf{79.61} & \textbf{49.06} \\
\hline
\end{tabular}%
}
\caption{Gender debiasing results of SEAT, StereoSet, and CrowSPairs on BERT, DistilBERT, and ELEATRA. The best result is indicated in \textbf{bold}. We report CCPA results with a continuous prompt template $(1,1,1)$. The closer the effect size is to 0 and the closer SS is to 50\%, the higher the fairness; the higher the LM and ICAT, the better. }
\label{tab:gender}
\end{table*}

\subsubsection{External Bias Evaluation Metrics}
\noindent\textbf{Bias-in-Bios} \citep{DBLP:conf/fat/De-ArteagaRWCBC19} is a biography dataset in which each sample is labeled with gender (male or female) and occupation (28 categories). 
We fine-tune the debiased model on the training set with the goal of predicting occupations. Overall Accuracy result is used to measure task precision, and individual Accuracy results for male and female are used to measure gender fairness. Furthermore, we report the gap between the true positive rates of the male prediction results and the female prediction results denotes as $GAP_{TPR}$, as well as the root mean square of the true positive rates difference for each category denotes as $GAP_{RMS}$. The closer their score is to 0, the better. They are defined as follows:
\begin{equation}
GAP_{TPR} = |TPR_M - TPR_F|,
\end{equation}
\begin{equation}
GAP_{RMS} = \sqrt{\frac{1}{|C|}\sum_{y\in C} (GAP_{TPR,y})^2}.
\end{equation}

\noindent\textbf{Bias-NLI} \citep{DBLP:conf/aaai/DevLPS20} fills gender words and occupation words with stereotypes into sentence templates to form sentence pairs, and the training goal is to inference whether the sentence pair is neutral or not. It defines three metrics to reflect the fairness of the model: 1) Net Neutral (NN), the average probability of neutral labels across all sentence pairs; 2) Fraction Neutral (FN), the proportion of sentence pairs marked as neutral; 3) Threshold:$\tau$ (T:$\tau$), The fraction of samples with neutral probability above $\tau$ is reported.

\subsection{Debiasing Performance Analysis}
\subsubsection{Internal Debiasing Results}
Table~\ref{tab:gender} shows the experimental results of three bias evaluation metrics for CCPA and baseline models on BERT, DistilBERT, and ELEATRA. We also report results for biased BERT, DistilBERT, and ELEATRA as references. The results show that CCPA achieves a better balance between PLMs' fairness and language modeling capability than the baseline models.

For BERT, CCPA reduces the average effect size from 0.621 to 0.249, increases ICAT from 66.86 to 73.28, and reduces CrowS from 57.86 to 51.57. Our method has achieved optimal results in the three test subsets of SEAT 6, 7, 8b and the average effect size, and has also been greatly improved in the other test subsets. The results on StereoSet show that CCPA does not weaken BERT's language modeling ability but slightly improves it. Although LM and SS do not achieve optimal results, our comprehensive index ICAT is better than other models. Both FairFil and MABEL are biased by contrastive learning, but their overall performance is not ideal. Although FairFil is outstanding in terms of SS performance, it seriously damages BERT's language modeling ability, possibly because it only considers sentence-level representation and does not retain token-level encoding ability. MABEL achieves promising results on StereoSet and CrowS-Pairs, but its SEAT results must be improved. Regarding overall performance, CCPA outperforms other contrastive learning frameworks, demonstrating that our adversarial training inspired approach can improve the model's learning ability by increasing the complex information in the model.

For DistilBERT, CCPA decreases the average effect size from 0.883 to 0.152 and improves ICAT from 66.93 to 71.30. Our model gets excellent experimental results on most test subsets of SEAT and reaches an almost ideal 50.31\% result on CrowS-Pairs. LM score decreases, and we analyze that the semantic information of the original representation is affected by too much debiasing.

For ELEATRA, which does not belong to the bert-series PLM, the debiasing effect of CCPA is equally significant, and the experimental results are fairer than the original ELEATRA on all three intrinsic metrics. In detail, CCPA reduced the average effect size from 0.797 to 0.421, increases ICAT by 8.37\% without significantly decreasing LM score, and reduces CrowS score by 1.89\%.

\begin{figure}[t]
\centering
\subfigure{
\begin{minipage}[t]{0.45\linewidth}
\includegraphics[width=1\linewidth]{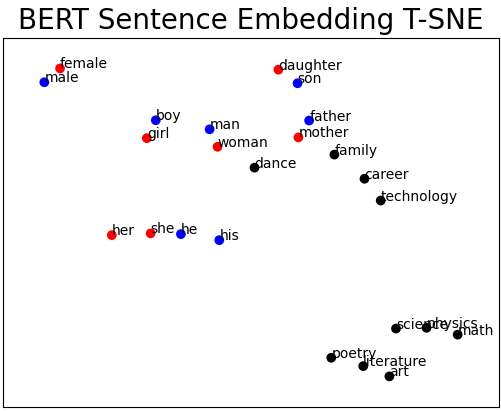}
\end{minipage}
}
\subfigure{
\begin{minipage}[t]{0.45\linewidth}
\includegraphics[width=1\linewidth]{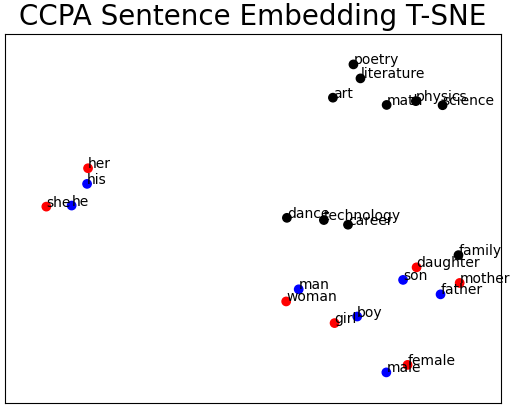}
\end{minipage}
}
\caption{T-SNE plots of sentence level representations encoded by BERT and CCPA. We use target words and their sentence templates in SEAT.}
\label{tsne}
\centering
\end{figure}

We also perform a small qualitative study by visualizing t-SNE plots of sentence embedding. As can be seen from Figure~\ref{tsne}, in BERT, male attribute words are more inclined to target words in the technical field (such as \textit{career} or \textit{science}) in the embedded space, while female attribute words are more inclined to target words in the humanities (such as \textit{family} or \textit{poetry}). After using CCPA to debias, it is observed that gender-attribute words are pulled closer together and away from neutral words in the representational space.

\begin{table}
\centering
\resizebox{0.48\textwidth}{!}{
\begin{tabular}{l|ccccc}
\hline
\multirow{2}*{Model}  & Acc. & Acc. & Acc. & GAP & GAP \\
                      & (All) &(M) & (F) & TPR & RMS \\
\hline
BERT & 84.14  &  84.69  &  83.50  &  1.189   &   0.144  \\
\hline
+INLP         & 70.50  &    -    &    -    &    -     &\textbf{0.067}  \\
+Sent-Debias  & 83.56  &  84.10  &  82.92  &  1.180   &   0.144  \\
+FairFil      & 83.18  &  83.52  &  82.78  &  0.746   &   0.142  \\
+MABEL        & \underline{84.85}  &  \underline{84.92}  &  \underline{84.34}  &  \underline{0.599}   &   0.132  \\
+CCPA (Ours)  & \textbf{85.65} & \textbf{85.41} & \textbf{85.95} &\textbf{0.544} & \underline{0.121} \\
\hline
\end{tabular}%
}
\caption{Results on Bias-in-Bios. The best result is indicated in \textbf{bold}. The sub-optimal result is indicated in \underline{underline}. '-' means not reported.}
\label{tab:bias-in-bios}
\end{table}

\begin{table}
\centering
\resizebox{0.45\textwidth}{!}{
\begin{tabular}{l|cccc}
\hline
Model  & NN & FN & T:0.5 & T:0.7 \\
\hline
BERT & 0.799  &  0.879  &  0.874  &  0.798   \\
\hline
+Sent-Debias  &  0.793  &  0.911  &  0.897  &  0.788  \\
+FairFil      &  0.829  &  0.883  &  0.846  &  0.845  \\
+MABEL        &  \textbf{0.900}  &  \textbf{0.977}  &  \textbf{0.974}  &  \textbf{0.935}  \\
+CCPA (Ours)  & \underline{0.883} & \underline{0.932} & \underline{0.929} & \underline{0.878} \\
\hline
\end{tabular}%
}
\caption{Results on Bias-NLI. The best result is indicated in \textbf{bold}. The sub-optimal result is indicated in \underline{underline}.}
\label{tab:bias-nli}
\end{table}

\subsubsection{External Debiasing Results}
We fine-tune the debiased BERT on two downstream tasks Bias-in-Bios and Bias-NLI to verify the effect of CCPA on external debiasing, and the results are shown in Tables \ref{tab:bias-in-bios} and \ref{tab:bias-nli}. All our experimental setups are consistent with MABEL, and all the results reported in the table for the baseline models are from MABEL. 

On the Bias-in-Bios task as shown in Table \ref{tab:bias-in-bios}, CCPA not only achieves the optimal results on task accuracy, but also performs the best on all gender fairness metrics except $GAP_{RMS}$. Although INLP obtains the best score on the $GAP_{RMS}$ metric, its task accuracy is clearly impaired from the reported results. Compared to all baselines, CCPA achieves the best overall debiasing performance while preserving the model's prediction performance on downstream tasks. 

On the Bias-NLI task as shown in Table \ref{tab:bias-nli}, CCPA achieves sub-optimal results on all the metrics. It is worth stating that MABEL is a debiasing method trained on the NLI task, which we analyze as the main reason for its most outstanding performance. Even so, the strong debiasing effect shown by CCPA on task Bias-NLI is heartening.  

The results of the internal debiasing experiment and the external debiasing experiment show that our proposed CCPA has outstanding performance in mitigating gender bias in PLMs' encoding. CCPA has an efficient debiasing performance, which answers the first question (\textbf{Q1}) proposed at the beginning of this section.

\begin{table*}
\centering
\resizebox{\textwidth}{!}{
\begin{tabular}{l|l|ccccccc|ccc|c}
\hline
\multicolumn{2}{c|}{BERT+} & SEAT-6 & SEAT-6b & SEAT-7 & SEAT-7b & SEAT-8 & SEAT-8b & Avg. & LM & SS & ICAT & CrowS \\
\hline
\multirow{3}*{$T_{(1, 1, 1)}$} & CCPA & 0.181 &-0.317& \textbf{0.104} & 0.633 & \textbf{0.142} & 0.115 &\textbf{0.249} & \textbf{84.44} &56.61 & \textbf{73.28} & 51.57 \\
& CCPA$^-$ & -0.198 & \textbf{-0.100} & -0.162 & \textbf{-0.309} & 0.535 & 0.243 & 0.258 & 79.34 & 57.31 & 67.75 & 47.17 \\
& CCPA$^*$ & \textbf{0.044} & -0.295 & -0.340 & 0.425 & -0.400 & \textbf{0.091} & 0.266 & 78.94 & \textbf{56.13} & 69.26 & \textbf{50.31} \\
\hline
\multirow{3}*{$T_{(2, 2, 2)}$} & CCPA & 0.126 & -0.135 & -0.379 & 0.144 & 0.416 & \textbf{0.056} & \textbf{0.209} & \textbf{82.40} & \textbf{55.36} & \textbf{73.56} & 51.57 \\
& CCPA$^-$ & 0.264 & \textbf{0.065} & \textbf{-0.372} & \textbf{0.127} & 0.150 & 0.481 & 0.243 & 79.23 & 57.44 & 67.44 & 48.43 \\
& CCPA$^*$ & \textbf{0.0918} & -0.178 & 0.509 & 0.311 & \textbf{0.144} & 0.271 & 0.251 & 79.84 & 54.94 & 71.95 & \textbf{50.94} \\
\hline
\multirow{3}*{$T_{(3, 3, 3)}$} & CCPA & \textbf{0.034} & \textbf{-0.006} & \textbf{0.193} & 0.278 & -0.189 & 0.346 & \textbf{0.174} & \textbf{82.62} & \textbf{54.80} & \textbf{74.68} & \textbf{49.06} \\
& CCPA$^-$ & 0.149 & 0.037 & -0.826 & \textbf{-0.106} & -0.124 & \textbf{-0.303} & 0.258 & 79.18 & 59.63 & 63.92 & 43.40 \\
& CCPA$^*$ & 0.119 & -0.131 & -0.334 & 0.225 & \textbf{-0.098} & -0.365 & 0.212 & 79.95 & 56.93 & 68.87 & \textbf{50.94} \\
\hline
\multicolumn{2}{l|}{NO$_{prompt}$} & 0.325 & 0.186 & 0.342 & 0.535 & 0.144 & 0.553 & 0.347 & 81.32 & 54.82 & 73.48 & 60.38 \\
\hline
\multicolumn{2}{l|}{NO$_{prompt+mask}$} & 0.730 & -0.012 & -0.185 & -0.530 & 0.927 & -0.158 & 0.424 & 61.85 & 53.98 & 56.92 & 34.59 \\
\hline
\end{tabular}%
}
\caption{Gender debiasing results of SEAT, StereoSet and CrowSPairs on BERT. The best result is indicated in \textbf{bold}. The closer the effect size is to 0 and the closer SS is to 50\%, the better; the higher the LM and ICAT, the better.}
\label{tab:ablation_bert}
\end{table*}

\subsection{Ablation Analysis}
We conduct ablation experiments on BERT to investigate how each component affects CCPA performance. The results are shown in Table~\ref{tab:ablation_bert}. 

$T_{(1, 1, 1)}$ indicates that the continuous prompt template is a triplet with one virtual token for each element, i.e., the length of prompts is 3. By analogy, $T_{(2, 2, 2)}$ and $T_{(3, 3, 3)}$ represent prompt templates of lengths 6 and 9. The purpose of this setting is to make it easier to observe the effect of the prompts' length on the model. In the experimental group of each template, we compare three versions of CCPA: the original CCPA, the version without $\mathcal{L}_{mlm}$ represented as CCPA$^-$ and the version without $\mathcal{L}_{mahal}$ represented as CCPA$^*$. In addition, we have experimented with both CCPA without prompts and CCPA without prompts and $\mathcal{L}_{mlm}$.

It is observed from the experimental results that the debiasing ability of CCPA increases with the rise of the template's length. This indicates that longer continuous prompt embeddings bring more difficult information to the model, thus increasing the debiasing effort. However, more extended templates can cause the original sentence semantics to be broken and thus weaken PLM's language modeling capability. In each experimental group, both CCPA$^-$ and CCPA$^*$ show a decrease in the results of the three evaluation metrics compared to CCPA. This phenomenon verifies that both MLM-assisted loss and Mahalanobis distance constraint benefit CCPA. Overall, MLM has a greater influence, especially on SS and CrowS, which may be because random mask tokens train encoders to retain token-level semantic information. 

In addition, the results of NO$_{prompt}$ verify that continuous prompts play an essential role in CCPA. NO$_{prompt+mask}$ tests the effect of fine-tuning PLMs based solely on contrastive learning. Unsurprisingly, the performance on all indexes could be better. The results of NO$_{prompt}$ and NO$_{prompt+mask}$ again reflect our method's effectiveness. The ablation studies answer our second question (\textbf{Q2}) by exploring the role played by each component of the CCPA.

\subsection{Language Modeling Capability Analysis}
We perform experiments on nine natural language understanding tasks of the GLUE benchmark to verify the language modeling capability of CCPA on downstream tasks. In task-specific fine-tuning, we set the learning rate to $2e-5$ and the batch size to 32 for all models.

As in Table~\ref{tab:glue}, CCPA's performance in 9 tasks is comparable to that of the original BERT, and the average results are almost equivalent to BERT's. CCPA also shows similar performance on DistilBERT, indicating that our model is effective on other models besides BERT. Combined with the LM score in Table~\ref{tab:gender}, the experiment shows that CCPA can debias without damaging the language modeling capability of PLMs, thus answering the third research question (\textbf{Q3}).

\section{Related Work}
We divide debiasing methods into two categories based on the debiasing strategy: task-specific methods and task-agnostic methods.

\subsection{Task-Specific Methods}
Task-specific methods adopt the strategy of debiasing in the fine-tuning stage of the downstream task, of which the downstream task is known \citep{DBLP:conf/eacl/HanBC21,DBLP:journals/corr/abs-2205-11485}. One representative work is INLP \citep{DBLP:conf/acl/RavfogelEGTG20,DBLP:conf/icml/RavfogelTGC22}, which repeatedly trains a linear classifier that predicts the target concept, and then projects the representation into the null space of the classifier's weight matrix to remove the representation bias. 
Contrastive learning is proposed to mitigate bias in classifier training~\citep{DBLP:journals/corr/abs-2109-10645}. It encourages instances sharing the same class labels to have similar representations while ensuring that protected attributes have different distributions. These methods use attribute words to label training data without CDA. However, they are biased towards specific downstream tasks and cannot be applied to other tasks in general. When training data change, task-specific methods are difficult to transfer to new tasks.

\begin{table*}
\small
\centering
\resizebox{\textwidth}{!}{
\begin{tabular}{l|cccccccccc}
\hline
Model &CoLA& MNLI& MRPC &QNLI& QQP& RTE& SST& STS-B &WNLI &Average \\
\hline
\textbf{BERT} &56.78 &84.76	&\textbf{89.54}	&91.51	&88.06	&64.62	&\textbf{93.35}	&88.24	&\textbf{56.34}	&79.24\\
+CDA &2.07	&84.84	&81.22	&84.84	&87.85	&47.29	&92.32	&40.83	&43.66	&62.77\\
+Dropout &2.07	&84.78	&81.22	&91.49	&88.02	&47.29	&92.09	&40.87	&43.66	&63.50 \\
+Sent-Debias &55.72	&\textbf{84.94}	&88.81	&91.54	&87.88	&63.90	&93.12	&88.23	&\textbf{56.34}	&78.94\\
+FairFil & 55.72 &84.85	&88.33	&\textbf{91.84}	&87.43	&64.98	&93.12	&88.55	&50.7 &78.39\\
+Auto-Debias &57.01	&84.91	&88.54	&91.65	&87.92 &64.62 &92.89 &88.43	&40.85	&77.42\\
+INLP &56.50 &84.78	&89.23	&91.38	&87.94	&\textbf{65.34}	&92.66	&88.73	&54.93	&79.05 \\
+MABEL &\textbf{57.80} &84.50 &85.00 &91.60 &\textbf{88.10} &64.30 &92.20 &\textbf{89.20} &- &\textbf{81.59} \\
+CCPA  & 55.91 & 84.73 & 88.65 & 91.42 & 87.98 & 64.93 &	93.09 & 88.44 &	55.66 & 78.98 \\
\hline
\textbf{DistilBERT}  & \textbf{47.93} &82.01 & \textbf{88.47} & \textbf{88.61} &86 68 & \textbf{58.84}	&90.71 &\textbf{86.26} &\textbf{56.34}	& \textbf{76.21} \\
+CCPA  &46.73 &	\textbf{82.53} &	86.99&	87.76&	\textbf{86.85} &	56.26&	\textbf{90.83} &	85.89&	55.93& 75.53\\
\hline
\end{tabular}%
}
\caption{Experimental results of GLUE tasks on BERT and DistilBERT. We report Matthew’s correlation for CoLA, the Spearman correlation for STS-B, and the F1 score for MRPC and QQP. Other tasks are reported for the accuracy. The best result is indicated in \textbf{bold}. '-' means not reported in MABEL.}
\label{tab:glue}
\end{table*}

\subsection{Task-Agnostic Methods}
Task-agnostic methods adopt the strategy of debiasing representation or processing unbalanced data before the downstream task, and they can be applied to any downstream task \citep{DBLP:conf/aaai/DevLPS20,DBLP:conf/emnlp/Dev0PS21}. Most of these methods apply counterfactual data augmentation to augment the unbalanced corpus and then debias the augmented text information. Counterfactual data augmentation \citep{DBLP:conf/birthday/LuMWAD20} is a general approach to augment corpora through causal intervention and has since been widely used to mitigate social biases. Different variants of counterfactual data augmentation have been proposed, such as Sent-Debias \citep{DBLP:conf/acl/LiangLZLSM20}, FairFil \citep{DBLP:conf/iclr/ChengHYSC21}, MABEL \citep{he2022mabel}, to name a few examples.

Task-agnostic methods primarily use the CDA to balance the training corpus by constructing counterfactual pairs specific to different demographic groups. However, simply applying CDA to the original corpus makes minor changes, constraining the representation space to a narrow range. This makes the model easily fit the differences between counterfactual pairs, weakening the debiasing ability.
Unlike existing CDA methods, we train a generator that encodes continuous prompts before fine-tuning PLM. The goal is to widen the representation distance between different groups to increase the difficulty of the model-learning process. 

\section{Conclusions}
Inspired by adversarial training, we propose CCPA, a two-stage debiasing model that combines contrastive learning with continuous prompts. In the continuous prompt tuning stage, we train a generator encoding continuous prompt embeddings to increase the representative distance between counterfactual pairs. In the fine-tuning stage, we use contrastive learning to reduce the representation distance between the augmented sample pairs. By increasing the difficulty of the training process, CCPA enables PLMs to learn a stronger debiasing ability. Extensive experiments on BERT and DistilBERT show that CCPA effectively reduces social bias in PLM representation while retaining language modeling capability.

\clearpage
\section*{Limitations}
In this work, we focus on debiasing the gender bias for PLMs.
In the future, we will try to mitigate social biases other than gender, such as race and religion. In addition, we also plan to extend our debiasing method to more language models, such as Natural Language Generation (NLG) models.

\section*{Ethics Statement}
This paper has been thoroughly reviewed for ethical considerations and has been found to be in compliance with all relevant ethical guidelines. The paper does not raise any ethical concerns and is a valuable contribution to the field.

\section*{Acknowledgments}
We express gratitude to the anonymous reviewers for their hard work and kind comments. The work was supported in part by the National Natural Science Foundation of China (No.62272191, No.61976102), the Science and Technology Development Program of Jilin Province (No.20220201153GX), the Interdisciplinary and Integrated Innovation of JLU (No.JLUXKJC2020207), and the Graduate Innovation Fund of Jilin University (No.2022214).

\bibliography{FairPrompts}
\bibliographystyle{acl_natbib}




\end{document}